\documentclass{article}

\usepackage[final]{corl_2026}           % camera-ready

\usepackage{amsmath,amssymb}
\usepackage{graphicx}
\usepackage{booktabs}
\usepackage{subcaption}
\usepackage{xcolor}
\usepackage{microtype}

\graphicspath{{assets/}}

\makeatletter
\renewcommand{\@noticestring}{%
  Workshop on \emph{Do Robots Need World Models?}, \@conferenceordinal\/ Conference on Robot Learning
  (CoRL \@conferenceyear), \@conferencelocation.}
\makeatother
\hypersetup{pdfauthor={Animesh Maheshwari},
  pdftitle={Prediction or Persistence? What Future-Prediction Scores Miss About World Models},
  pdfsubject={Workshop on Do Robots Need World Models? at CoRL 2026}}

\DeclareMathOperator*{\argmin}{arg\,min}
\newcommand{\anchor}{A}                 % anchor (context) length, seconds
\newcommand{\rollout}{W}                % rollout (horizon) length, seconds
\newcommand{\pers}{\mathrm{pers}}

\title{Prediction or Persistence?\\What Future-Prediction Scores Miss\\About World Models}

\author{
  Animesh Maheshwari\\
  Deccan AI
}

\begin{document}
\maketitle

%===============================================================================
\begin{abstract}
Pretrained video world models are compared and chosen as components for robot learning largely by future-prediction scores. On its own, such a score cannot be read as evidence of prediction, because it mixes encoder quality, the near future's resemblance to the recent past, and the predictor's own contribution. We separate the three with a matched control. On $27{,}000$ future-retrieval trials from household egocentric video, a model must pick the true continuation of a clip among eleven candidates from the same video (chance $0.091$), ranked zero-shot by its own training criterion. Each model is paired with a \emph{persistence control}, an identical copy whose predictor is never invoked, so the paired difference isolates the predictor. Score and contribution dissociate within the native-video V-JEPA family alone: the predictor adds $+0.291$ top-1 for V-JEPA~2.1 but $+0.03$--$0.04$ for V-JEPA~2 on the identical trials. Across five models the contribution ranges from $+0.291$ to negative, and it is negative only for the two action-conditioned models, which we evaluate outside their training regime. Persistence alone recovers the encoder ordering expected from pretraining, a sign that the probe tracks real signal. We suggest reporting each future-prediction score together with its persistence baseline. The gap between them, not the score, is what makes the number interpretable: a necessary check costing one forward pass. Establishing downstream usefulness is a separate matter, for a planning or control evaluation.
\end{abstract}


%===============================================================================
\section{Introduction}
\label{sec:intro}

Robot learning increasingly builds on pretrained video world models, such as V-JEPA~2, DINO-WM and the wider latent-predictive line, as components a policy is initialised from, plans against, or is evaluated with~\citep{ha2018worldmodels,lecun2022path,assran2025vjepa2,zhou2025dinowm}. Which one to trust is decided largely by future-prediction score. If that score misleads, the decisions built on it (which model to pretrain from, plan against, or evaluate with) inherit the error. Our probe uses passive video with no robot in the loop: it concerns what the score measures, not control. That is still a question for robot learning, because the model under test and the decision the score informs are the same ones.

A future-prediction score reflects three things at once: an encoder that represents the scene well, a near future that resembles the recent past, and a predictor that anticipates what comes next. Only the last is prediction in the intended sense, yet the score sums all three. A predictor that contributes nothing still scores well on the first two, and a harmful one can still beat chance. The score alone therefore cannot show what the predictive machinery adds.

\begin{figure}[t]
  \centering
  \includegraphics[width=0.58\textwidth]{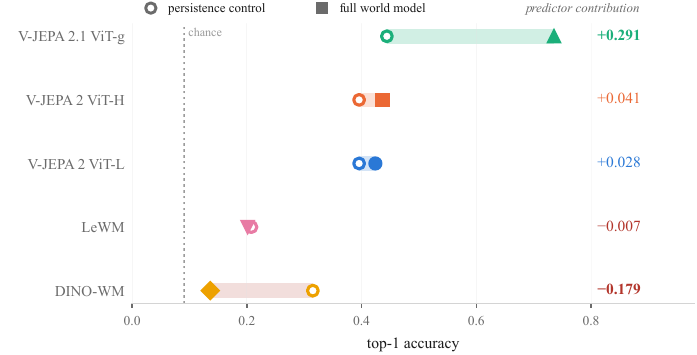}
  \caption{\textbf{What the predictor adds.} Top-1 on the same $27{,}000$ trials for each model (filled) and its own persistence control (open), an identical copy whose predictor is never called; the shaded span is the paired contribution $\delta_\theta$. Score and contribution are close to unrelated: V-JEPA~2 ViT-H and its control both sit near $0.4$. DINO-WM and LeWM are measured off-distribution, so their contributions are illustrative.}
  \label{fig:contribution}
\end{figure}

We separate the three with a matched control. A model must identify a clip's true continuation among eleven candidates from the \emph{same} video. It is paired with a \textbf{persistence control}, an identical copy whose predictor is never invoked, isolating the predictor for one extra forward pass. Score and contribution diverge (Fig.~\ref{fig:contribution}). Within the native-video V-JEPA family, V-JEPA~2's predictors add $+0.028$ to $+0.041$ top-1 to scores above $0.42$, while V-JEPA~2.1's adds $+0.291$ on the identical trials. DINO-WM, measured off-distribution, scores $0.136$ against $0.091$ chance, yet its persistence baseline reaches $0.315$. We therefore offer a diagnostic for reading the score: without the persistence gap, a future-prediction number cannot be interpreted. We do not try to settle whether world models are needed; we show that the score alone cannot settle it, and we measure the predictor's share directly. \textbf{Contributions:} (i)~a persistence control decomposing any future-prediction score into encoder quality, temporal persistence and predictor contribution, for one forward pass and no training; (ii)~a $27{,}000$-trial zero-shot future-retrieval probe over $1{,}094$ egocentric videos, sweeping context and horizon independently; (iii)~predictor contributions for five models from $+0.291$ to negative, pooled negatives confined to the two off-distribution models, and a positional envelope fixing in advance where a frozen predictor leaves its trained range.

%===============================================================================

%===============================================================================
\section{Related Work}
\label{sec:related}

\textbf{Latent-predictive world models} predict in representation space: V-JEPA pairs a masked-feature-prediction video encoder with a latent predictor~\citep{assran2023ijepa,bardes2024vjepa}, scaled in V-JEPA~2 to over a million hours of video~\citep{assran2025vjepa2} and retrained for dense features in V-JEPA~2.1~\citep{murlabadia2026vjepa21}. DINO-WM learns an action-conditioned predictor over frozen DINOv2 features~\citep{oquab2024dinov2,zhou2025dinowm}, and LeWorldModel learns both from pixels~\citep{maes2026lewm}. Reconstruction-based models~\citep{hafner2025dreamerv3,finn2016unsupervised} and video simulators~\citep{bruce2024genie,nvidia2025cosmos} are compared by downstream success. Headline scores have repeatedly become interpretable only once paired with a \textbf{trivial baseline}: constant velocity for pedestrians~\citep{scholler2020constantvelocity}, tuned metric-learning baselines~\citep{musgrave2020reality}, copying the present in anticipation~\citep{girdhar2021avt,han2019dpc}. World-model comparisons are mostly reported without one. Our persistence control is the world-model analogue: the \emph{same model} with one module disabled, so the difference is attributable.

%===============================================================================
\section{Method}
\label{sec:method}

The diagnostic has two parts: a retrieval probe that produces a future-prediction score, and a matched control that removes the predictor from it.

\paragraph{Future retrieval.} A trial is $\tau=(V,t,\anchor,\rollout)$: the model observes the \emph{anchor} $X_\tau=V[t-\anchor,\,t]$ and must identify the ground-truth future $y^\star_\tau=V[t,\,t+\rollout]$ among $N=11$ candidates, each a $\rollout$-second window of the \emph{same} video. Scene recognition is therefore uninformative, and chance top-1 is $1/N=0.0909$. With frozen encoder $f_\theta$, frozen predictor $g_\theta$ and native training criterion $E_\theta$ (lower is more compatible),
\begin{equation}
  \hat\jmath_\tau \;=\; \argmin_{j\in\{1,\dots,N\}} \;
      E_\theta\!\big(\,\underbrace{g_\theta(f_\theta(X_\tau),\,\rollout)}_{\text{predicted future}},\; f_\theta(c_j)\big).
  \label{eq:decision}
\end{equation}
Distractors tile \emph{outward} from the anchor's start and the ground truth's end, so none overlaps the anchor. Presentation order and the number preceding it are redrawn per trial, so neither position nor geometry reveals the answer. Each model is scored by its own training criterion over candidates embedded by its \emph{own} encoder, since an energy is only meaningful in the space its objective was defined over. We compare only the induced ranking (Appendices~\ref{app:setup},~\ref{app:energy}; worked trials in Fig.~\ref{fig:trial}).

\paragraph{The persistence control.} The control asks whether the explicit predictor adds anything over an encoder that copies the present, that is, whether the predictive structure is doing any work. Egocentric video is temporally continuous, so much of any score needs no prediction. We pair every model with a control identical except that the predictor is replaced by the identity, $g^{\pers}_\theta(z,\rollout)=z$: it asks ``which candidate most resembles what was just observed?'', under the same encoder, preprocessing and energy. The \textbf{predictor contribution} is the paired difference in top-1 over a common trial set $\mathcal{T}$,
\begin{equation}
  \delta_\theta \;=\; \frac{1}{|\mathcal{T}|}\sum_{\tau\in\mathcal{T}}
     \Big[\mathbf{1}\{\hat\jmath_\tau=j^\star_\tau\} - \mathbf{1}\{\hat\jmath^{\,\pers}_\tau=j^\star_\tau\}\Big].
  \label{eq:delta}
\end{equation}
It attributes to the predictor alone, since encoder, energy and trials are held fixed, and is signed: $\delta_\theta<0$ marks a predictor worse than not predicting. Retrieval among candidates tests temporal localisation rather than the content of a predicted future, so a positive $\delta_\theta$ certifies that the predictor carries future-relevant information beyond its encoder: a prerequisite for, not a demonstration of, usefulness in planning or control. Because trials cluster in videos, every top-1 and $\delta_\theta$ interval is a $95\%$ \textbf{video-level cluster bootstrap} over the $704$ videos ($B=10{,}000$, percentile, seeded; Appendix~\ref{app:cluster}).

\paragraph{Corpus, envelope and protocol.} Trials are built from $1{,}094$ household egocentric videos ($479$\,h, no action stream), every window inside a span of annotated manipulation. We sweep $\anchor\in\{1,\dots,10\}$\,s and $\rollout\in\{1,1.5,\dots,\anchor\}$\,s over $100$ cells, separating input budget from horizon. A frozen predictor can only be queried at positions it was trained over. The anticipation wrapper lays context and target tokens end to end, so they consume the positional budget additively: $\anchor+\rollout\le uP/\phi$. For V-JEPA~2.1 ViT-g ($P{=}32$, $u{=}2$, $\phi{=}8$\,fps) that bound is $8$\,s. We evaluate inside that envelope and report outside it separately. Every cross-model number uses one pinned intersection of $\mathbf{27{,}000}$ trials ($704$ videos, $25$ cells) scored by all nine variants at their own published geometries.

%===============================================================================
\begin{figure}[t]
  \centering
  \includegraphics[width=0.66\textwidth]{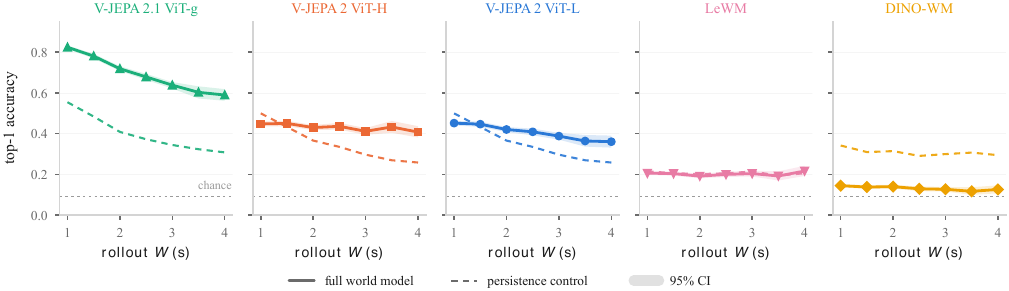}
  \caption{\textbf{Top-1 against prediction horizon $\rollout$}, model (solid) versus its own persistence control (dashed). V-JEPA~2.1 decays in parallel with persistence; V-JEPA~2's curves are flat and \emph{cross} their control at $\rollout{=}1$\,s; DINO-WM, measured off-distribution, sits below its control everywhere. Bands are $95\%$ trial-level: too few videos per cell to cluster-bootstrap, so only Table~\ref{tab:headline}'s pooled statistics are cluster-corrected (Appendix~\ref{app:cluster}).}
  \label{fig:horizon}
\end{figure}

%===============================================================================
\section{Results}
\label{sec:results}

The central finding is that score and predictor contribution dissociate. Within the native-video V-JEPA family alone, scores above $0.42$ coexist with contributions of $+0.03$ and $+0.04$, while V-JEPA~2.1's predictor adds $+0.291$, before any off-distribution model is considered.

\paragraph{The contribution is not read off the score.} The dissociation holds within the native-video V-JEPA family and survives dropping V-JEPA~2.1 entirely. V-JEPA~2 ViT-H and ViT-L reach $0.437$ and $0.424$ top-1, over four times chance, against a shared control at $0.396$. Their predictors therefore contribute only $+0.0408$ $[+0.0326,\,+0.0491]$ and $+0.0284$ $[+0.0205,\,+0.0363]$. The claim that \emph{a high score need not mean the predictor contributes} therefore rests on neither the single 2.1 checkpoint nor any action-conditioned model. V-JEPA~2.1 ViT-g (Fig.~\ref{fig:contribution}) widens the spread: $0.736$ against $0.444$, $\delta=+0.291$ $[+0.281,\,+0.302]$, roughly $10\times$ V-JEPA~2's. Neither persistence row calls a predictor, so the $+0.340$ between the two families splits into $+0.048$ encoder and $+0.291$ predictor (Table~\ref{tab:decomp}), terms a single accuracy cannot separate. Native energy is the stricter readout: it needs no trained probe, so a small or negative $\delta_\theta$ reflects the predictor under its own training objective rather than a readout we chose.

\paragraph{A predictor can be worse than none.} DINO-WM scores $0.136$, above the $0.091$ chance rate, which on its own would suggest a working predictor, while its persistence control scores $0.315$. Removing the predictor improves retrieval, so scoring above chance was not evidence that it predicts. Its predictor expects actions in PushT's action space, which passive household video cannot supply, so it is measured off-distribution: we report the direction and treat the magnitude, $-0.179$ $[-0.186,\,-0.172]$, as illustrative (see Limitations). The same direction appears in-distribution below, where V-JEPA~2 falls under its own control at $\rollout{=}1$\,s with no embodiment gap. LeWM is off-distribution for the same reason and serves as a domain-mismatch control rather than a peer.

\paragraph{Persistence alone recovers the pretraining ordering.} With every predictor disabled, the four encoders order video- $>$ image- $>$ sim-pretrained, with all four intervals disjoint. Persistence top-1 is $0.444$ for ViT-g, $0.396$ for ViT-L, $0.315$ for DINOv2 and $0.208$ for ViT-Ti, trained from scratch in simulation.

\paragraph{Horizon separates anticipating from copying.} Persistence decays with the rollout $\rollout$ ($0.554\!\to\!0.308$ for the ViT-g encoder, $1$--$4$\,s) as the future drifts from the present. V-JEPA~2.1 decays in parallel ($0.824\!\to\!0.590$), holding $+0.27$ to $+0.31$ throughout, so its predictor buys a roughly constant horizon (Fig.~\ref{fig:horizon}). V-JEPA~2's curves are nearly flat ($0.452\!\to\!0.361$ for ViT-L) and sit \emph{below} their control at $\rollout{=}1$\,s, $0.452$ against $0.500$. At short horizons its predictor is worse than copying. This happens inside the native-video family with no embodiment gap, so it is a counterpart to DINO-WM without that confound.

%===============================================================================
\section{Discussion}
\label{sec:discussion}

\paragraph{A reporting norm.} We suggest that \emph{every future-prediction score be reported alongside its own persistence baseline}. It costs one forward pass, needs no training, and applies to any encoder--predictor factorisation. Without it, the reported number cannot show whether the predictive structure is doing any work; with it, that share is measured directly. On our passive-video probe the share is substantial for V-JEPA~2.1 and marginal for V-JEPA~2, and the negative values for the two off-distribution models are illustrative. This does not settle whether robots need world models, but it adds a comparison that is usually missing from that discussion.

%===============================================================================
\clearpage
\section*{Limitations}

\paragraph{Passive-video scope.} The probe scores \emph{predictive representation} on passive household video. No claim about control, planning or downstream robot performance follows. Retrieval among eleven candidates tests temporal localisation rather than the content of a predicted future, and it is easier than generation.

\paragraph{Action inputs.} DINO-WM and LeWM are action-conditioned models trained on simulated PushT, and passive video supplies no actions in that space. We evaluate them with zero actions through their own action encoders, so their predictors run outside the regime they were built for. Their persistence controls never call the predictor and are unaffected, but their full-model figures are confounded, so their magnitudes are illustrative and the thesis rests on the V-JEPA models.

\paragraph{Other confounds.} The anchor axis is confounded by lineup geometry and is not interpreted; the corpus is a single household domain; only one V-JEPA~2.1 checkpoint is evaluable, so that result carries no scaling curve. Clustering is not among the confounds: its largest widening ($2.12\times$) falls on V-JEPA~2.1's top-1, and every claim still holds (Appendix~\ref{app:cluster}). Appendix~\ref{app:limits} gives the full discussion and the experiments that would remove each limitation.

\section*{AI Usage Statement}
The author used generative AI tools for writing and editing assistance. All technical content, experiments, and claims are the author's own.

\bibliography{main}

%===============================================================================
\clearpage
\appendix

\section{Benchmark construction, models and inference}
\label{app:setup}

\paragraph{Corpus.} $1{,}094$ household egocentric videos, $478.6$\,h, $242{,}356$ annotated subtasks, passive first-person capture with no action stream. An \emph{active run} is a maximal contiguous span of annotated manipulation, merging gaps of $\le 1.0$\,s; there are $25{,}296$ such runs covering $425.8$\,h ($89\%$ of the corpus). All lineup geometry must fit inside one active run, so every window shows genuine manipulation rather than idle or transition footage. A trial is labelled by the action in the \emph{ground-truth} window, not the anchor, so the label describes what has to be predicted.

\paragraph{Lineup geometry.} With margin $m$ and stride $s=\rollout+m$, and with $k$ distractors drawn per trial to precede the anchor (at least one on each side), the candidate set is
\begin{equation}
  \resizebox{\textwidth}{!}{$\displaystyle
  \mathcal{C}_\tau =
  \underbrace{\big\{[\,t{-}\anchor{-}m{-}(i{-}1)s{-}\rollout,\; t{-}\anchor{-}m{-}(i{-}1)s\,]\big\}_{i=1}^{k}}_{\text{before the anchor}}
  \;\cup\; \underbrace{[\,t,\;t{+}\rollout\,]}_{y^\star_\tau}
  \;\cup\; \underbrace{\big\{[\,t{+}\rollout{+}m{+}(i{-}1)s,\; t{+}2\rollout{+}m{+}(i{-}1)s\,]\big\}_{i=1}^{N-1-k}}_{\text{after the ground truth}},$}
  \label{eq:lineup}
\end{equation}
presented in a per-trial seeded permutation. We use $m=0$, giving a contiguous strip. A validator enforces three properties and rejects any lineup that violates them: no distractor overlaps the anchor; presentation order is randomised; and the answer's index is not a fixed function of the geometry, because $k$ is redrawn per trial. Verified on $40{,}000$ sampled lineups: $0\%$ anchor-overlapping distractors, $0\%$ in raw time order, ground-truth index approximately uniform over all $11$ slots. Further construction parameters: anchor stride $0.5$\,s, minimum anchor separation $2.0$\,s, at most $3$ datapoints per (video, cell), ground truth never at a temporal edge, annotator-confidence floor $0.8$, global seed $0$.

\paragraph{From construction to the evaluated set.} Construction produced $282{,}032$ lineups; balancing to $400$ trials per (task group, cell) retained $204{,}715$ (surplus is marked, not deleted). Of these, $54{,}481$ satisfy $\anchor+\rollout\le 8$\,s, and $27{,}000$ were scored by all nine model variants and pinned by id. The $8$\,s restriction is principled (\S\ref{sec:method}); the further restriction to $27{,}000$ is a wall-clock budget, roughly $2\times$ fewer than could be scored with no new construction. The evaluated set spans $704$ videos and $25$ cells with $1{,}041$--$1{,}131$ trials per cell (median $1{,}081$); composition is in Table~\ref{tab:composition}.

\paragraph{Models and checkpoints.} Each model runs at its own published evaluation geometry: V-JEPA~2 ViT-L/16 ($300$\,M, $256$\,px, $1024$-d) and ViT-H/16 ($600$\,M, $256$\,px); V-JEPA~2.1 ViT-g/16 ($1$\,B, $384$\,px, $4{\times}1408{=}5632$-d hierarchical target)~\citep{assran2025vjepa2,murlabadia2026vjepa21,vjepa21release}; DINO-WM ($21$\,M frozen DINOv2 ViT-S/14 encoder $+$ $20$\,M predictor, $224$\,px, $196{\times}384$-d) trained on simulated PushT~\citep{zhou2025dinowm}; LeWM ($15$\,M; $\sim$$5$\,M ViT-Ti/14 encoder \emph{trained from scratch}, $224$\,px, $192$-d), also on simulated PushT~\citep{maes2026lewm}. Two selection points are material to reproducibility. (i) V-JEPA~2 and~2.1 are separate families with different modules, encoder keys and input resolutions. (ii) The released 2.1 ViT-B and ViT-L are \emph{distilled}: their predictors emit their ViT-G teacher's $1664$-d features, so their native energy is undefined without also running the teacher. ViT-g is used because it is trained with the 2.1 recipe directly and its predictor targets its own EMA encoder. This also means the 2.1 result rests on a single checkpoint, with no scaling curve inside that family.

\paragraph{Action conditioning.} DINO-WM and LeWM are action-conditioned control models, and the corpus is passive video. Actions and proprioception are therefore supplied as \emph{zeros} through the models' own action/proprio encoders rather than injected post-embedding, because those layers have biases. Their predictors are conditioned on PushT's action space, which passive household video cannot supply and which no video-derived surrogate would occupy either, so both models are measured off-distribution here. The mismatch is confined to one arm of each pair: an action is an input to the predictor, not to the encoder, so a control that never calls the predictor never receives an action. In both adapters the zero action is in fact constructed inside the roll-forward method that the control overrides, so the persistence figures are action-free by construction. That makes the control side of each pair clean; it does not make the full-model figure unconfounded, which is why \S\ref{sec:results} reports those magnitudes as illustrative.

\paragraph{Inference.} Decode once per video at $8$\,fps and share the frames across models; bfloat16, except LeWM whose \texttt{encode()} casts its input and therefore runs in float32; token-budgeted batching; $4{\times}$A100 80\,GB, one worker per GPU over a largest-first video-bundle queue. Clip length follows the window's duration rather than being resampled to a fixed frame count, so one tubelet corresponds to the same physical time in every cell and a horizon in seconds converts to tubelet steps exactly. Top-1 is primary; mean rank and R@$k$ accompany it; intervals are $95\%$ video-level cluster bootstraps (Appendix~\ref{app:cluster}), the per-trial normal approximation being retained only as a cross-check on the machinery; model-vs-control differences use the paired form of Eq.~\eqref{eq:delta}. All sampling, shuffling and bootstrapping is seeded ($\texttt{seed}=0$).

\section{Native energy}
\label{app:energy}

For a prediction $\hat z\in\mathbb{R}^{T\times d}$ and candidate tokens $h\in\mathbb{R}^{T\times d}$, the V-JEPA pretraining objective is a LayerNormed $\ell_p$ regression,
\begin{equation}
  E_\theta(\hat z,h) \;=\; \frac{1}{Td}\sum_{\iota=1}^{T}\big\|\hat z_\iota - \mathrm{LN}(h_\iota)\big\|_p^p\big/p,\qquad p=1 .
  \label{eq:energy}
\end{equation}
The LayerNorm must be kept, since omitting it compares the prediction against a quantity the model was never trained to match. The V-JEPA~2.1 predictor emits the concatenated features of $B{=}4$ encoder blocks, and its training loss normalises \emph{each block separately} before concatenation, $\mathrm{LN}_B(h)=[\mathrm{LN}(h^{(1)})\|\cdots\|\mathrm{LN}(h^{(B)})]$; a single LayerNorm over the full $Bd$ width would mix blocks of different scale into one statistic and is not the quantity the predictor was trained against. DINO-WM is scored by its \texttt{emb\_criterion} (MSE over the visual token slice) and LeWM by its \texttt{criterion} (MSE summed over features). In every case lower energy denotes greater compatibility.

\paragraph{The degeneracy check.} Each trial is additionally re-ranked using the ground-truth candidate's \emph{own} latent as the query. Since that query \emph{is} a member of $\mathcal{C}_\tau$ its minimum is a self-distance and it ranks first by construction; the check therefore detects an encoder that collapses distinct moments onto identical latents. It attains $1.000$ for every model reported here and should not be read as an achievable-accuracy bound.

\section{Additional results}
\label{app:figures}

\begin{figure}[h]
  \centering
  \includegraphics[width=\textwidth]{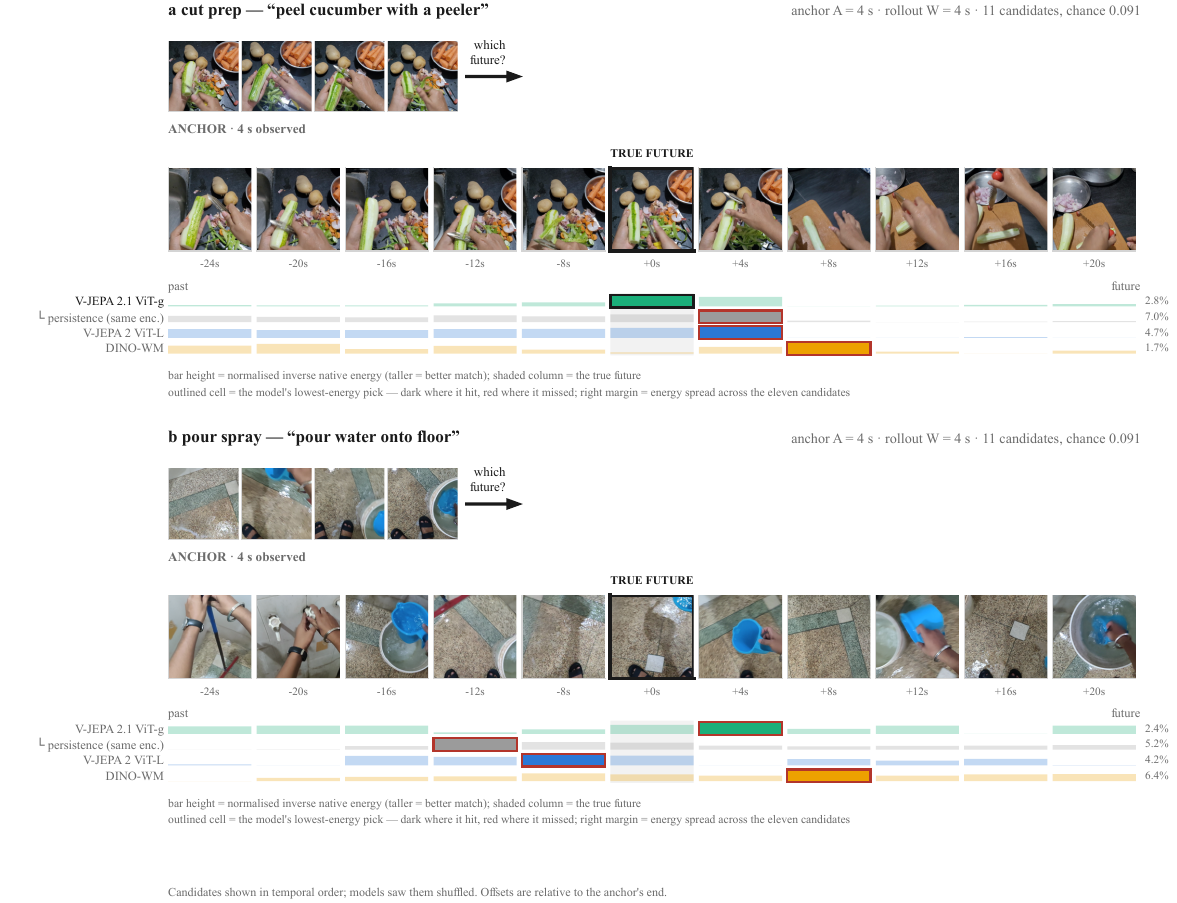}
  \caption{\textbf{Two worked trials.} Each panel shows a $4$\,s anchor followed by the eleven $4$\,s candidates drawn from the same video, in temporal order (models see them shuffled), with offsets relative to the anchor's end; the shaded column is the true future. Bars are normalised inverse native energy (taller is a better match) and the outlined cell is each model's lowest-energy pick, dark where it hit and red where it missed; the right margin gives the energy spread across the eleven candidates. \textbf{(a)} V-JEPA~2.1 recovers the true continuation, while its own persistence control (same frozen encoder, predictor never called) slides one step later, and DINO-WM (measured off-distribution) lands two steps past it. \textbf{(b)} A harder trial in which every model misses, the near misses falling on candidates adjacent to the truth.}
  \label{fig:trial}
\end{figure}

\begin{figure}[h]
  \centering
  \begin{subfigure}[b]{0.47\textwidth}
    \centering\includegraphics[width=\textwidth]{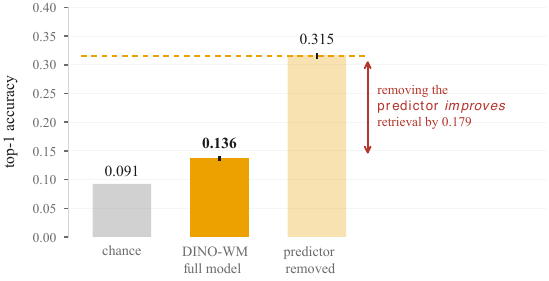}
    \caption{A predictor that costs accuracy}
  \end{subfigure}\hfill
  \begin{subfigure}[b]{0.51\textwidth}
    \centering\includegraphics[width=\textwidth]{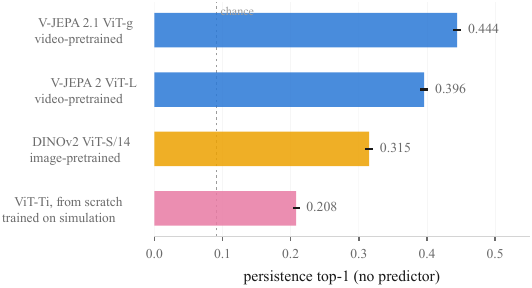}
    \caption{Encoder ordering with every predictor off}
  \end{subfigure}
  \caption{\textbf{Two readings that only the control makes available.} (a) DINO-WM is above chance, yet deleting its predictor raises top-1 from $0.136$ to $0.315$; it is measured off-distribution here, so the magnitude is illustrative (\S\ref{sec:results}). (b) With every predictor disabled, persistence top-1 orders the four encoders video-pretrained $>$ image-pretrained $>$ sim-from-scratch, a validity check the benchmark was not tuned for.}
  \label{fig:cases}
\end{figure}

\begin{figure}[h]
  \centering
  \includegraphics[width=\textwidth]{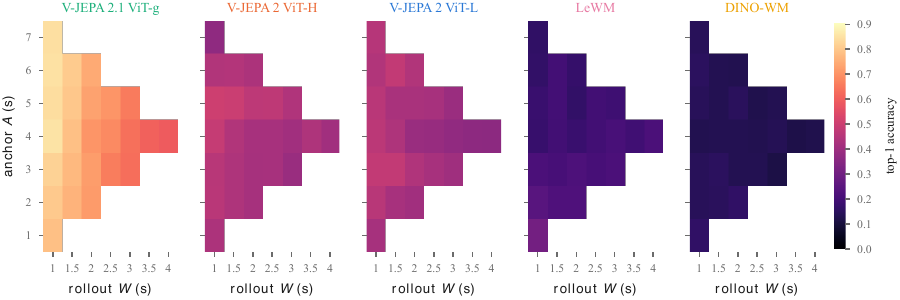}
  \caption{Top-1 for every evaluated $(\anchor,\rollout)$ cell. Accuracy falls along the rollout axis $\rollout$ for all five models. The apparent rise along the anchor axis $\anchor$ is a geometric artefact of the lineup construction and is not interpreted (Table~\ref{tab:byA}).}
  \label{fig:grid}
\end{figure}

\begin{figure}[h]
  \centering
  \includegraphics[width=\textwidth]{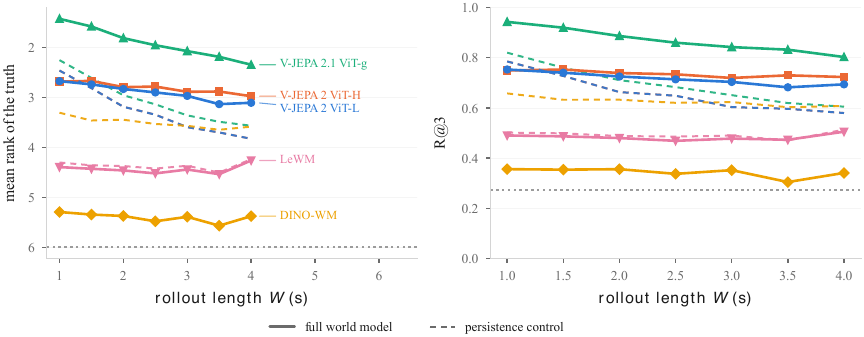}
  \caption{Mean rank of the ground truth (left) and R@3 (right) against $\rollout$; solid is the full model, dashed its persistence control. Every ordering and every model-vs-control sign matches the top-1 result, so the conclusions are not an artefact of thresholding at rank~1.}
  \label{fig:rank}
\end{figure}

\begin{figure}[h]
  \centering
  \includegraphics[width=0.6\textwidth]{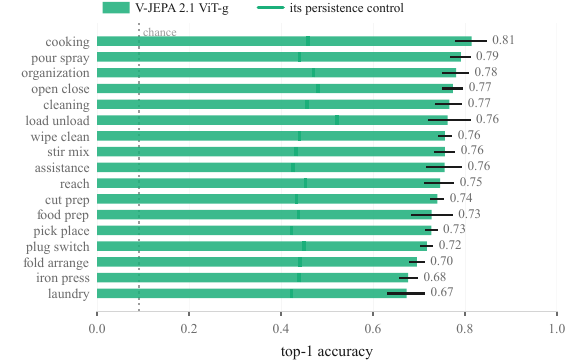}
  \caption{V-JEPA~2.1 ViT-g by task group (bars, $95\%$ trial-level: per-group video counts are too small to cluster-bootstrap, so only the pooled statistics of Table~\ref{tab:headline} are cluster-corrected; Appendix~\ref{app:cluster}) against its persistence control (vertical rule). Top-1 ranges from $0.67$ (laundry) to $0.81$ (cooking), and all $17$ groups above the $1\%$ threshold (Table~\ref{tab:composition}) sit above the control, so no task family is carried by persistence alone.}
  \label{fig:tasks}
\end{figure}

\paragraph{Error structure.} $0.264$ of top-1 is unreached by the strongest model, and its mistakes are not systematic drift. Candidates are ordered in time, so a signed index offset says whether a model that misses reaches into the past (negative) or overshoots (positive). The mean signed offset is $-0.02$ steps with a $50/50$ past/future split for every model, and only $18\%$ of errors land on a candidate adjacent to the truth. All three are indistinguishable from a uniform guess ($-0.02$; $0.502/0.498$; $0.182$). Failures are therefore losses of temporal localisation rather than a bias toward regressing to the past, so closing the remaining gap is not a matter of sharper sub-second timing.

\begin{figure}[h]
  \centering
  \includegraphics[width=\textwidth]{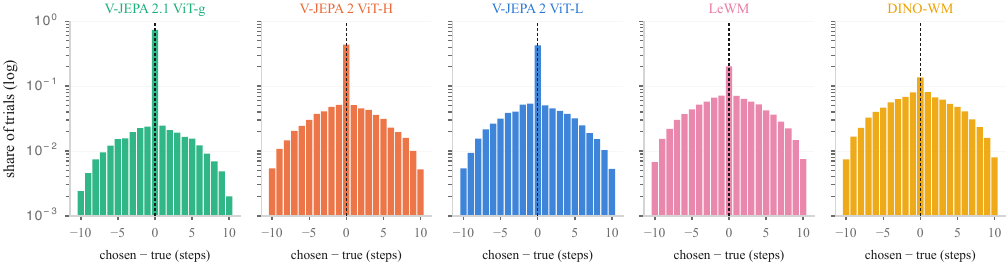}
  \caption{Distribution of (chosen $-$ true) candidate index, log scale. The spike at $0$ is the accuracy; the surrounding mass is broad and near-symmetric for every model.}
  \label{fig:confusion}
\end{figure}

\begin{figure}[h]
  \centering
  \includegraphics[width=0.6\textwidth]{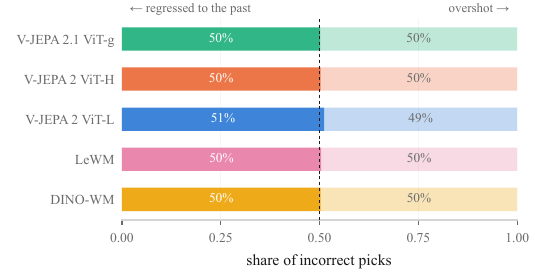}
  \caption{Direction of incorrect picks. Every model splits $50/50$ between reaching into the past and overshooting into the future, matching a uniform guess. No model, including those whose predictors hurt, shows a systematic bias toward copying the past.}
  \label{fig:direction}
\end{figure}

\clearpage
\section{Full tables}
\label{app:tables}

\begin{table}[h]
  \centering\small
  \caption{\textbf{Decomposing V-JEPA~2.1 ViT-g's $0.736$.} Both persistence rows call no predictor, so the gain from the V-JEPA~2 recipe to the 2.1 recipe splits into a representation term and a prediction term. The last row is exact retrieval, \emph{not} achievable headroom.}
  \label{tab:decomp}
  \resizebox{\textwidth}{!}{%
  \begin{tabular}{lcl}
    \toprule
    \textbf{level} & \textbf{top-1} & \textbf{increment attributable to} \\
    \midrule
    chance ($1/11$)                        & $0.091$ & --- \\
    persistence, V-JEPA~2 ViT-L encoder    & $0.396$ & $+0.305$\quad video continuity $+$ representation \\
    persistence, V-JEPA~2.1 ViT-g encoder  & $0.444$ & $+0.048$\quad a better encoder, still zero prediction \\
    \textbf{V-JEPA~2.1 ViT-g (full model)} & $\mathbf{0.736}$ & $\mathbf{+0.291}$\quad \textbf{the predictor} \\
    perfect retrieval                      & $1.000$ & $+0.264$\quad unsolved \\
    \bottomrule
  \end{tabular}}
\end{table}

\begin{table}[h]
  \centering\small
  \caption{Headline results: all nine model variants on the identical $27{,}000$ trials. Rows marked \textcolor{gray}{$\llcorner$} are persistence controls, which share the frozen encoder of the row above and never call a predictor. All intervals are $95\%$ video-level cluster bootstraps over the $704$ videos (Appendix~\ref{app:cluster}); $\delta_\theta$ is the paired contribution and every one excludes zero.}
  \label{tab:headline}
  \resizebox{\textwidth}{!}{%
  \begin{tabular}{lcccccc}
    \toprule
    \textbf{model} & \textbf{top-1} & \textbf{cluster 95\% CI} & \textbf{mean rank} & \textbf{R@3} & \textbf{R@5} & $\boldsymbol{\delta_\theta}$ \textbf{(cluster 95\% CI)} \\
    \midrule
    V-JEPA 2.1 ViT-g/16                & $0.736$ & $[0.7243,\,0.7466]$ & $1.75$ & $0.894$ & $0.946$ & $\mathbf{+0.2914}$ $[+0.2806,\,+0.3019]$ \\
    \textcolor{gray}{$\llcorner$} Persistence (ViT-g enc.)  & $0.444$ & $[0.4359,\,0.4523]$ & $2.81$ & $0.732$ & $0.853$ & --- \\
    V-JEPA 2 ViT-H/16                  & $0.437$ & $[0.4291,\,0.4441]$ & $2.76$ & $0.740$ & $0.864$ & $\mathbf{+0.0408}$ $[+0.0326,\,+0.0491]$ \\
    V-JEPA 2 ViT-L/16                  & $0.424$ & $[0.4161,\,0.4321]$ & $2.82$ & $0.729$ & $0.858$ & $\mathbf{+0.0284}$ $[+0.0205,\,+0.0363]$ \\
    \textcolor{gray}{$\llcorner$} Persistence (ViT-L enc.)  & $0.396$ & $[0.3879,\,0.4036]$ & $3.03$ & $0.694$ & $0.833$ & --- \\
    DINO-WM (DINOv2 ViT-S/14)          & $0.136$ & $[0.1319,\,0.1408]$ & $5.37$ & $0.351$ & $0.537$ & $\mathbf{-0.1789}$ $[-0.1861,\,-0.1718]$ \\
    \textcolor{gray}{$\llcorner$} Persistence (DINOv2 enc.) & $0.315$ & $[0.3084,\,0.3222]$ & $3.45$ & $0.635$ & $0.787$ & --- \\
    LeWM (ViT-Ti/14, from scratch)     & $0.201$ & $[0.1961,\,0.2068]$ & $4.44$ & $0.484$ & $0.668$ & $\mathbf{-0.0066}$ $[-0.0098,\,-0.0034]$ \\
    \textcolor{gray}{$\llcorner$} Persistence (LeWM enc.)   & $0.208$ & $[0.2026,\,0.2134]$ & $4.36$ & $0.494$ & $0.679$ & --- \\
    \midrule
    \textit{chance}                    & $0.091$ & ---        & $6.00$ & $0.273$ & $0.455$ & --- \\
    \bottomrule
  \end{tabular}}
\end{table}

\begin{table}[h]
  \centering\small
  \caption{Top-1 by rollout length $\rollout$ (pooled over $\anchor$). \texttt{pers-}$x$ is the persistence control built from encoder $x$: $g$ = V-JEPA~2.1 ViT-g, $L$ = V-JEPA~2 ViT-L, Dv2 = DINOv2 ViT-S/14, Ti = LeWM ViT-Ti. Note that \texttt{pers-}$L$ \emph{exceeds} both V-JEPA~2 models at $\rollout{=}1$\,s.}
  \label{tab:byW}
  \resizebox{\textwidth}{!}{%
  \begin{tabular}{lccccccccc c}
    \toprule
    $\rollout$ (s) & 2.1 ViT-g & pers-$g$ & 2 ViT-H & 2 ViT-L & pers-$L$ & DINO-WM & pers-Dv2 & LeWM & pers-Ti & $n$ \\
    \midrule
    $1.0$ & $0.824$ & $0.554$ & $0.448$ & $0.452$ & $0.500$ & $0.144$ & $0.342$ & $0.206$ & $0.215$ & $7{,}510$ \\
    $1.5$ & $0.781$ & $0.484$ & $0.450$ & $0.447$ & $0.433$ & $0.138$ & $0.309$ & $0.204$ & $0.206$ & $5{,}312$ \\
    $2.0$ & $0.719$ & $0.409$ & $0.430$ & $0.420$ & $0.366$ & $0.140$ & $0.314$ & $0.191$ & $0.202$ & $5{,}383$ \\
    $2.5$ & $0.679$ & $0.373$ & $0.436$ & $0.409$ & $0.335$ & $0.129$ & $0.291$ & $0.200$ & $0.205$ & $3{,}298$ \\
    $3.0$ & $0.638$ & $0.344$ & $0.411$ & $0.388$ & $0.297$ & $0.127$ & $0.300$ & $0.204$ & $0.214$ & $3{,}305$ \\
    $3.5$ & $0.603$ & $0.324$ & $0.433$ & $0.364$ & $0.269$ & $0.117$ & $0.307$ & $0.192$ & $0.194$ & $1{,}088$ \\
    $4.0$ & $0.590$ & $0.308$ & $0.408$ & $0.361$ & $0.258$ & $0.126$ & $0.294$ & $0.216$ & $0.206$ & $1{,}104$ \\
    \bottomrule
  \end{tabular}}
\end{table}

\begin{table}[h]
  \centering\small
  \caption{Top-1 by anchor length $\anchor$ (pooled over $\rollout$). \textbf{This axis is confounded} and is included only for completeness: distractors tile backwards from the anchor's start, so a longer anchor pushes them further from the ground truth and makes them easier to reject. Nothing about \emph{does more context help} is answerable from this table.}
  \label{tab:byA}
  \resizebox{\textwidth}{!}{%
  \begin{tabular}{lccccccccc c}
    \toprule
    $\anchor$ (s) & 2.1 ViT-g & pers-$g$ & 2 ViT-H & 2 ViT-L & pers-$L$ & DINO-WM & pers-Dv2 & LeWM & pers-Ti & $n$ \\
    \midrule
    $1$ & $0.781$ & $0.380$ & $0.432$ & $0.416$ & $0.369$ & $0.160$ & $0.402$ & $0.304$ & $0.327$ & $1{,}077$ \\
    $2$ & $0.755$ & $0.428$ & $0.436$ & $0.427$ & $0.393$ & $0.150$ & $0.355$ & $0.231$ & $0.239$ & $3{,}147$ \\
    $3$ & $0.716$ & $0.411$ & $0.423$ & $0.442$ & $0.376$ & $0.135$ & $0.311$ & $0.214$ & $0.219$ & $5{,}414$ \\
    $4$ & $0.691$ & $0.407$ & $0.430$ & $0.396$ & $0.353$ & $0.128$ & $0.306$ & $0.195$ & $0.201$ & $7{,}620$ \\
    $5$ & $0.742$ & $0.472$ & $0.473$ & $0.423$ & $0.414$ & $0.135$ & $0.296$ & $0.184$ & $0.189$ & $5{,}452$ \\
    $6$ & $0.796$ & $0.519$ & $0.438$ & $0.453$ & $0.459$ & $0.137$ & $0.310$ & $0.177$ & $0.183$ & $3{,}209$ \\
    $7$ & $0.841$ & $0.619$ & $0.368$ & $0.452$ & $0.556$ & $0.143$ & $0.315$ & $0.161$ & $0.161$ & $1{,}081$ \\
    \bottomrule
  \end{tabular}}
\end{table}

\begin{table}[h]
  \centering\small
  \caption{Predictor contribution $\delta_\theta$ inside and outside the positional envelope, measured on the widest set each model covered (the V-JEPA~2 family spans all $100$ cells, so $n$ is larger than the matched $27{,}000$). The first three rows lie inside the $\anchor+\rollout\le 8$\,s envelope. Spans beyond it are an \emph{out-of-distribution measurement}, obtained only by widening the predictor's positional address space at inference, and must not be read as a capability limit. These spans fall outside the matched evaluated set, so their intervals are per-trial normal approximations rather than cluster bootstraps.}
  \label{tab:envelope}
  \begin{tabular}{lccc}
    \toprule
    span $\anchor+\rollout$ & V-JEPA 2 ViT-L & V-JEPA 2 ViT-H & V-JEPA 2.1 ViT-g \\
    \midrule
    $0$--$4$\,s   & $+0.037\pm0.011$ & $+0.035\pm0.011$ & $+0.330\pm0.015$ \\
    $4$--$6$\,s   & $+0.032\pm0.008$ & $+0.037\pm0.009$ & $+0.290\pm0.012$ \\
    $6$--$8$\,s   & $+0.036\pm0.007$ & $+0.053\pm0.007$ & $+0.277\pm0.010$ \\
    \midrule
    $8$--$11$\,s  & $+0.022\pm0.016$ & $-0.018\pm0.017$ & $+0.052\pm0.050$ \\
    $11$--$14$\,s & $+0.027\pm0.014$ & $-0.051\pm0.014$ & $-0.137\pm0.047$ \\
    $14$--$21$\,s & $+0.060\pm0.013$ & $-0.006\pm0.013$ & $-0.074\pm0.040$ \\
    \bottomrule
  \end{tabular}
\end{table}

\begin{table}[h]
  \centering\small
  \caption{Composition of the evaluated $27{,}000$ trials; a trial is labelled by the action in its ground-truth window. Left: all nine tasks. Right: the $17$ task groups above the $1\%$ reporting threshold, as shown in Fig.~\ref{fig:tasks}; six further groups (dress, wring\_twist, plant\_care, hosting, home\_maintenance, daily\_routine) account for the remaining $2.9\%$. A task group is a family of action verbs; a verb outside every family falls back to its video's task, so the groups marked $^\dagger$ share a task's name but hold only that task's unmapped actions (e.g.\ most laundry trials fall under fold\_arrange and iron\_press).}
  \label{tab:composition}
    \begin{tabular}{lcc@{\hskip 2.5em}lcc}
    \toprule
    \textbf{task} & \textbf{trials} & \textbf{share} & \textbf{task group} & \textbf{trials} & \textbf{share} \\
    \midrule
    laundry           & $5{,}860$ & $21.7\%$ & plug\_switch  & $4{,}050$ & $15.0\%$ \\
    organization      & $5{,}780$ & $21.4\%$ & pick\_place   & $3{,}671$ & $13.6\%$ \\
    cleaning          & $5{,}461$ & $20.2\%$ & cut\_prep     & $2{,}907$ & $10.8\%$ \\
    food\_prep        & $3{,}426$ & $12.7\%$ & wipe\_clean   & $2{,}826$ & $10.5\%$ \\
    cooking           & $3{,}079$ & $11.4\%$ & fold\_arrange & $2{,}820$ & $10.4\%$ \\
    assistance        & $2{,}618$ & $ 9.7\%$ & iron\_press   & $1{,}788$ & $ 6.6\%$ \\
    daily\_routine    & $361$     & $ 1.3\%$ & stir\_mix     & $1{,}300$ & $ 4.8\%$ \\
    hosting           & $305$     & $ 1.1\%$ & open\_close   & $1{,}287$ & $ 4.8\%$ \\
    home\_maintenance & $110$     & $ 0.4\%$ & pour\_spray   & $1{,}273$ & $ 4.7\%$ \\
                      &           &          & organization$^\dagger$ & $817$ & $ 3.0\%$ \\
                      &           &          & cleaning$^\dagger$     & $745$ & $ 2.8\%$ \\
                      &           &          & reach         & $677$     & $ 2.5\%$ \\
                      &           &          & cooking$^\dagger$      & $486$ & $ 1.8\%$ \\
                      &           &          & laundry$^\dagger$      & $447$ & $ 1.7\%$ \\
                      &           &          & assistance$^\dagger$   & $446$ & $ 1.7\%$ \\
                      &           &          & food\_prep$^\dagger$   & $349$ & $ 1.3\%$ \\
                      &           &          & load\_unload  & $320$     & $ 1.2\%$ \\
    \midrule
    \textbf{total}    & $\mathbf{27{,}000}$ & $\mathbf{100\%}$ & & & \\
    \bottomrule
  \end{tabular}
\end{table}

\clearpage
\section{Cluster-robust confidence intervals}
\label{app:cluster}

The $27{,}000$ evaluated trials are drawn from $704$ videos and $4{,}425$ active runs, so roughly $38$ trials share each video. Trials from one video share footage, scene, lighting and actor, so their outcomes are correlated and the effective sample size is below $27{,}000$; a per-trial normal-approximation interval is therefore too narrow.

\paragraph{Method.} A video-level cluster bootstrap over the evaluated set. For each of $B=10{,}000$ replicates we draw $704$ video ids with replacement from the $704$ unique ones, gather every trial belonging to the drawn videos (a video drawn twice contributes its trials twice), and recompute the statistic; the $2.5$th and $97.5$th percentiles of the replicate distribution are the interval. Three properties matter. \emph{Resampling is at the video level, never the trial level}: all trials from a chosen video stay together, which is what carries the dependence. \emph{Pairing is preserved}: for a contribution $\delta_\theta$ the per-trial indicator of Eq.~\eqref{eq:delta} is formed first, then videos are resampled once per replicate and the difference read from that same sample, so a model and its control are never resampled independently. \emph{Point estimates are untouched}: they come from the full data and only the intervals change. No re-inference is involved: per-trial correctness is read from the stored results. Seeded, so it reproduces exactly.

\paragraph{Is $B$ large enough?} Re-running under $8$ seeds gives the Monte-Carlo error of the procedure itself, which is unrelated to the statistical uncertainty being measured. Table~\ref{tab:breps} reports how far the interval endpoints move between seeds. The intervals themselves are $0.0064$--$0.0223$ wide; at $B=10{,}000$ the endpoint jitter is $\pm0.00029$, about $9.1\%$ of the narrowest interval's half-width, so it affects only the fourth decimal place. $B=10{,}000$ is more than sufficient, and even $2{,}000$ would be defensible. Bootstrap bias, $|\text{mean of replicates} - \text{point estimate}|$, is $\le 0.00007$ across all $14$ statistics, so plain percentile intervals are appropriate and a BCa or studentised bootstrap would change nothing.

\begin{table}[h]
  \centering\small
  \caption{Endpoint jitter across $8$ seeds at each $B$: the Monte-Carlo error of the bootstrap itself, not the uncertainty it estimates. The final column is the worst of all $14$ reported statistics (nine top-1 values and five contributions).}
  \label{tab:breps}
  \resizebox{\textwidth}{!}{%
  \begin{tabular}{lcccccc c}
    \toprule
    replicates $B$ & 2.1 ViT-g top-1 & 2 ViT-L top-1 & DINO-WM top-1 & $\delta$ 2.1 ViT-g & $\delta$ 2 ViT-L & $\delta$ DINO-WM & worst of $14$ \\
    \midrule
    $500$      & $\pm0.00083$ & $\pm0.00095$ & $\pm0.00061$ & $\pm0.00124$ & $\pm0.00104$ & $\pm0.00059$ & $\pm0.00124$ \\
    $1{,}000$  & $\pm0.00078$ & $\pm0.00064$ & $\pm0.00038$ & $\pm0.00086$ & $\pm0.00064$ & $\pm0.00047$ & $\pm0.00086$ \\
    $2{,}000$  & $\pm0.00067$ & $\pm0.00029$ & $\pm0.00020$ & $\pm0.00064$ & $\pm0.00033$ & $\pm0.00046$ & $\pm0.00067$ \\
    $5{,}000$  & $\pm0.00050$ & $\pm0.00028$ & $\pm0.00014$ & $\pm0.00032$ & $\pm0.00014$ & $\pm0.00023$ & $\pm0.00050$ \\
    $\mathbf{10{,}000}$ & $\pm0.00029$ & $\pm0.00026$ & $\pm0.00008$ & $\pm0.00020$ & $\pm0.00014$ & $\pm0.00018$ & $\pm0.00029$ \\
    $20{,}000$ & $\pm0.00020$ & $\pm0.00008$ & $\pm0.00007$ & $\pm0.00014$ & $\pm0.00012$ & $\pm0.00009$ & $\pm0.00020$ \\
    \bottomrule
  \end{tabular}}
\end{table}

\paragraph{Is the video the right grain?} Trials ($27{,}000$) nest in active runs ($4{,}425$), runs nest in videos ($704$), and videos nest in actors ($352$); every video is its own episode, so there is no hidden level between video and actor. Resampling at each grain (Table~\ref{tab:grain}) separates real dependence from over-correction. Widening from trial to run to video is dependence being captured, up to $2.12\times$. Widening from video to actor is not: at most $1.13\times$, and $1.00\times$ on the headline contribution. Video level therefore absorbs nearly all the clustering, and resampling more coarsely only widens the intervals. Every $\delta_\theta$ still excludes zero even at actor level. As a check on the machinery, the trial-level bootstrap reproduces the analytic normal-approximation interval almost exactly: $\pm0.0053$ against $\pm0.0053$ for V-JEPA~2.1 top-1.

\begin{table}[h]
  \centering\small
  \caption{Half-widths under resampling at each grain. Video level (bold) is what we report.}
  \label{tab:grain}
  \begin{tabular}{lcccc}
    \toprule
    statistic & trial ($n{=}27{,}000$) & run ($4{,}425$) & \textbf{video} ($704$) & actor ($352$) \\
    \midrule
    2.1 ViT-g top-1        & $\pm0.0053$ & $\pm0.0078$ & $\mathbf{\pm0.0112}$ & $\pm0.0119$ \\
    2 ViT-L top-1          & $\pm0.0060$ & $\pm0.0069$ & $\mathbf{\pm0.0080}$ & $\pm0.0088$ \\
    DINO-WM top-1          & $\pm0.0041$ & $\pm0.0043$ & $\mathbf{\pm0.0045}$ & $\pm0.0048$ \\
    $\delta$ 2.1 ViT-g     & $\pm0.0066$ & $\pm0.0082$ & $\mathbf{\pm0.0107}$ & $\pm0.0106$ \\
    $\delta$ 2 ViT-L       & $\pm0.0068$ & $\pm0.0072$ & $\mathbf{\pm0.0079}$ & $\pm0.0082$ \\
    $\delta$ DINO-WM       & $\pm0.0064$ & $\pm0.0068$ & $\mathbf{\pm0.0072}$ & $\pm0.0072$ \\
    \bottomrule
  \end{tabular}
\end{table}

\paragraph{Outcome, and reproduction.} No point estimate changes, no estimate changes sign, and no contribution interval crosses zero; the four persistence intervals remain disjoint, so the video- $>$ image- $>$ sim-pretrained encoder ordering holds under clustering. The intervals in Table~\ref{tab:headline} are the cluster intervals. The correction applies to those pooled statistics only: bands in the per-cell and per-task-group figures (Figs.~\ref{fig:horizon},~\ref{fig:tasks}) remain trial-level, because within a single $(\anchor,\rollout)$ cell or task group too few videos contribute to resample reliably, so those bands show spread rather than cluster-robust uncertainty. The bootstrap ($B{=}10{,}000$, $\texttt{seed}=0$) runs in ${\sim}35$\,s from the per-trial scores. Code, the pinned trial ids and the per-trial scores for every model will be released publicly.

\clearpage
\section{Extended limitations and the experiments that remove them}
\label{app:limits}

\paragraph{Construct validity.} The probe measures predictive representation on passive video, so no claim about control, planning or model-predictive use follows. Retrieval among $11$ candidates is strictly easier than generation: a model can rank correctly while its predicted latent is far from the target in absolute terms, and scores here upper-bound nothing about rollout fidelity. Zero-shot native-energy retrieval is a protocol choice; the official V-JEPA evaluation trains a 4-block attentive probe, which measures something related but looser. The degeneracy check is not headroom (Appendix~\ref{app:energy}).

\paragraph{Confounds in the model comparison.} Model families differ on several axes at once (resolution, latent width, pretraining corpus, predictor objective, parameter count), so the persistence controls isolate the predictor's contribution \emph{within} a family but cannot attribute the V-JEPA~2\,$\to$\,2.1 gap to any single change. Only one V-JEPA~2.1 checkpoint is evaluable, so that result carries no scaling curve. LeWM is a domain-mismatch control, not a peer of the video-pretrained models. DINO-WM and LeWM are additionally measured outside the action-conditioned regime they were built for, which is why the thesis is carried by the V-JEPA contributions and their magnitudes are reported as illustrative.

\paragraph{Ranked next steps.} Ordered by priority.
\begin{enumerate}\itemsep2pt
  \item \textbf{De-confound the anchor axis.} Place distractors at \emph{fixed temporal offsets from the ground truth} rather than tiling from the anchor's start, so varying $\anchor$ changes only the amount of context and leaves candidate geometry identical. Table~\ref{tab:byA} would then become interpretable, and the design could measure how much context a predictive model actually uses.
  \item \textbf{Scale to the full in-range pool.} The evaluated set is budget-limited, not construction-limited: roughly twice as many in-range lineups already exist and need no new data preparation, which would tighten the small-$n$ task groups.
  \item \textbf{Relate to the official probe protocol.} Run the standard 4-block attentive probe alongside native-energy retrieval on the same trials. This anchors our numbers to the published V-JEPA results, and the gap between protocols is itself informative: it measures how much of a model's usable future information is inaccessible without a trained readout.
  \item \textbf{Extend past the 8\,s bound.} In increasing cost: positional interpolation or RoPE scaling at inference; a sliding-context formulation that keeps every query inside the trained range; or continued pretraining at longer context. Any of these would turn the out-of-envelope rows of Table~\ref{tab:envelope} into a tested result about long-horizon prediction.
  \item \textbf{Broaden the corpus, and add an action-conditioned split.} A second domain tests generalisation. Separately, the only fair evaluation of an action-conditioned world model is the identical future-retrieval lineup built over a robot dataset with \emph{native recorded actions}: manufacturing surrogate actions cannot escape the embodiment mismatch, so we leave this to future work.
  \item \textbf{Stratify by difficulty, and add a human baseline.} Label each trial by scene change, motion magnitude and semantic distance between the target and its nearest distractor, then report accuracy by stratum, separating temporal-precision failures from semantic ones. A human baseline on a few hundred trials per stratum would establish whether the $0.264$ of top-1 that the strongest model misses is headroom or irreducible ambiguity. This is the most important quantity the study does not yet measure.
\end{enumerate}

\paragraph{Reproduction.} Every trial row stores the full $11$-candidate energy vector, so any re-ranking metric can be recomputed without re-running inference. The evaluated set is pinned by id, and all sampling, shuffling and bootstrapping is seeded, so the figures reproduce exactly.

\end{document}